# BubGAN: Bubble Generative Adversarial Networks for Synthesizing Realistic Bubbly Flow Images

**Yucheng Fu, Yang Liu**

Nuclear Engineering Program, Mechanical Engineering Department, Virginia Tech
635 Prices Fork Road, Blacksburg, VA 24061
ycfu@vt.edu; liu130@vt.edu

**ABSTRACT**

Bubble segmentation and size detection algorithms have been developed in recent years for their high efficiency and accuracy in measuring bubbly two-phase flows. In this work, we proposed an architecture called bubble generative adversarial networks (BubGAN) for the generation of realistic synthetic images which could be further used as training or benchmarking data for the development of advanced image processing algorithms. The BubGAN is trained initially on a labeled bubble dataset consisting of ten thousand images. By learning the distribution of these bubbles, the BubGAN can generate more realistic bubbles compared to the conventional models used in the literature. The trained BubGAN is conditioned on bubble feature parameters and has full control of bubble properties in terms of aspect ratio, rotation angle, circularity and edge ratio. A million bubble dataset is pre-generated using the trained BubGAN. One can then assemble realistic bubbly flow images using this dataset and associated image processing tool. These images contain detailed bubble information, therefore do not require additional manual labeling. This is more useful compared with the conventional GAN which generates images without labeling information. The tool could be used to provide benchmarking and training data for existing image processing algorithms and to guide the future development of bubble detecting algorithms.

## 1 Introduction

Dense object detecting and counting is a common but time-consuming and challenging task. The applications can be found in the field of pedestrian surveillance [1], vehicle detection in aerial images [2], cell or bacterial colony counting in medical images [3–5], oil droplet characterization in petroleum engineering [6] and bubble counting in bubble columns or nuclear reactors. In bubble column or nuclear reactor applications, accurate separation and reconstruction of bubble shape have equal importance to the number counting, since bubble shape contains important geometrical information for the study of mass, momentum and energy transport in these systems [6,7]. The demand for bubble shape acquisition introduces additional challenges for algorithm development and benchmarking in this field.

High speed imaging is a powerful technique which can be used to record bubbly flow images at high spatial and temporal resolutions. The image processing technique can be used to extract bubble parameters from these images. Recently, the reported image processing algorithms [8–20] can deal with complicated conditions such as high void fraction flows, severe bubble deformation and overlapping, etc. These algorithms can reduce the cost of processing bubbly flow images considerably and provide detailed information about bubble size distribution, shape, volume, etc. One issue in developing these algorithms is the way to benchmark the accuracy of these algorithms. Currently, the benchmark strategies can be divided into two categories. The first one is to compare the algorithm with different measurement techniques, which include conductivity probe [21,22], x-ray [23], or global gas and liquid flow meters. The benchmark with global instruments provides an overall error estimation without local uncertainty information. The comparison with conductivity probe or x-ray method can be made for time- or line-averaged parameters. The other issue in this benchmarking strategy is that these measurement methods may contain large uncertainty and can may not be used to assess image processing algorithms [22].

The second strategy is to use synthetic images as the benchmarking data [14,24]. With the automation of image synthesizing process, the cost and time for the benchmarking of image processing algorithm can be reduced. The primary error source for this method comes from the gap between the real bubbly flow images and the synthetic ones. Current algorithms for synthetic images are mostly limited to generating simple bubble shapes such as spherical or elliptical bubbles. These physical models usually assume the bubble edge intensity follows a concentric circular/elliptical arrangements (CCA) [25,26]. With given bubble size and distribution information, synthetic bubbly flow images can be generated for benchmarking purpose. However, these algorithms are not capable of modeling fine structures of bubble shapes and intensity variations. Synthetic bubbles of different size can have a similar intensity distribution. Therefore, image processing algorithms benchmarked with these synthetic images can have a quite different performance in processing real bubbly flow images.

Conventional image processing algorithms are mostly based on certain features of bubble images such as curvature, intensity gradient, and topology information. The feature design and selection require the expertise in the related field to achieve an optimized performance. The deep learning algorithms such as convolutional neural networks (CNN), which do not require the input of extracted features from images, can be an alternative solution for bubble detection [26,27]. The CNN can process bubbly flow image in their raw form without pre-processing for number density detection. For those supervised deep learning algorithms, labeled images are required for algorithm training. Presently, the researches in using deep learning algorithm for bubbly recognition and separation in the chemical and nuclear engineering fields are limited. One of the bottlenecks is the lack of a large amount of high-quality labeled data for algorithm training. In other fields, using synthetic data becomes a trend for training deep networks to reduce the manual labeling cost and to improve the efficiency. However, this can be effective only if we can generate synthetic data which is close to the real world [28].

To bridge the gap between real bubbly flow images and synthetic ones, we proposed a new approach called bubble generative adversarial networks (BubGAN) for generating realistic bubbly flow images. Conventional GAN algorithms can generate realistic images but they cannot be directly used in this research. The bubbly flow images directly generated with GAN have no label information of the bubbles. In addition, the resolution of the generated images with GAN is relatively low compared to the recorded high-resolution images from high speed cameras. To overcome these shortcomings, the proposed BubGAN algorithm adopts the "divide and conquer strategy" to achieve high resolution labeled image generation. The BubGAN combines the conventional image processing algorithms [14,17] and the generative adversarial networks [29,30] for realistic bubble synthesis. In the algorithm, the GAN will be only responsible for single bubble generation rather than generating the bubbly flow images directly. The image processing algorithm is responsible for GAN training data preparation and bubble assembling for bubbly flow image synthesis. With given bubbly flow boundary conditions, the synthetic bubbly flow images can be generated by assembling single synthetic bubbles on an image background canvas. In these images, bubble location, edge boundary, rotation, etc., of all bubbles will be labeled for either existing algorithm benchmarking or development of new algorithms.

The structure of this paper is organized as follows. In section 2, the experimental setup for acquiring bubbly flow images is described. High-speed images are recorded in an upward rectangular two-phase flow test loop. In section 3, the methodologies used in the BubGAN are introduced. A million-bubble database is generated for bubbly flow image synthesis with the BubGAN algorithm. Section 4 presents both qualitative and quantitative studies of the BubGAN results.

## 2 Experimental setup

In this study, the experimental bubbly flow images are recorded by a high-speed camera in a rectangular channel. The schematic of the test facility is shown in Fig. 1 (a). The rectangular channel is designed for adiabatic air-water two-phase upflow at room temperature and atmospheric pressure. The 3.0 m tall test section features a 30 mm × 10 mm rectangular cross section. Compressed air injected into the test section is measured by four gas flow meters based on the laminar differential pressure flow technology, which has an accuracy of about ±1% of the actual reading. The water flow rate entering the test section is measured by two magnetic flow meters with an accuracy of about ±1% of the actual reading. The design of the two-phase injector at the bottom of the test channel is shown in Fig. 1 (b). For the water flow injected at the bottom, a flow straightener (honeycomb) is used to provide a uniform water injection and reduce the turbulence. Two aluminum plates (in orange) with miniature holes are installed flush on the two 30-mm-wide walls for air injection.

There are three instrumentation ports along the test section. These ports can be used to house flow measurement instruments or used for flow visualization. In between these instrumentation ports are three visualization sections with a larger height for continuous two-phase flow structure visualization. An example of the flow visualization section is shown in Fig. 3 (c). The hydraulic diameter of this test section is $D_h$ = 15 mm. In this study, the bubbly flow images are recorded at the height of $z/D_h$ = 8.8 with a high speed camera (Photron FASTCAM SA4). The high speed camera is mounted to face the 30 mm-wide side of the test section. A LED panel is placed opposite to the high speed camera for illumination. Images were recorded at 1000 frame per second (fps) and have a resolution of around 25 pixels per mm. The superficial gas velocity is fixed at $j_{g0}$ = 0.1 m/s and the superficial liquid velocity is set as $j_f$ = 0.5 m/s. The superficial velocity is defined as the volume flow rate of the liquid/gas phase divided by the cross section area.

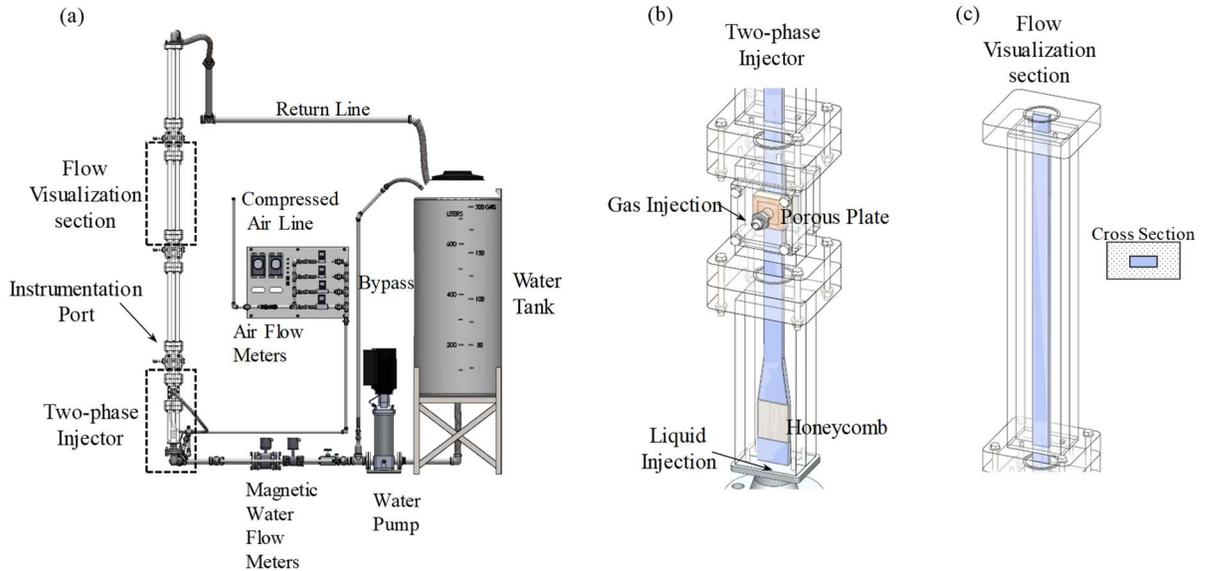

Fig. 1 (a) Schematic of the experimental facility for running upward air-water two-phase bubbly flow, (d) an enlarged view of the two-phase injector at the bottom of the rectangular test channel, (c) an enlarged view of an upper rectangular section for continuous two-phase flow structure visualization.

## 3 Methodology

This section presents the methodologies used in the BubGAN for the generation of synthetic bubbly flow images. As shown in Fig. 2, the BubGAN algorithm consists of three major steps. The first step is to process the recorded high-speed images to extract single bubble images. The second step trains a conditional GAN to generate synthetic bubble images based on the training data acquired from the previous step. This deep convolutional generative adversarial neural network (DCGAN) is conditioned on four selected bubble features, which are of primary interest in bubbly flow study. This enables the users to generate a specific bubble with given parameters for different flow conditions. To reduce the runtime cost, a million-bubble database is pre-generated based on the trained conditional GAN. With given bubble size distribution and other related physical information, synthetic bubbles can be directly sampled from the million-bubble database and be assembled to generate synthetic bubbly flow images. The tool for synthetic bubbly flow image generation is available at https://github.com/ycfu/BubGAN.

In these steps, two primary techniques, namely, image processing and conditional GAN are used in the BubGAN algorithm. As shown in Fig. 2, the processes associated with image processing techniques are shown in red, and those with conditional GAN in purple. The blocks marked in blue represent the physical information required for bubbly flow image generation, and the yellow blocks represent the data used in the BubGAN.

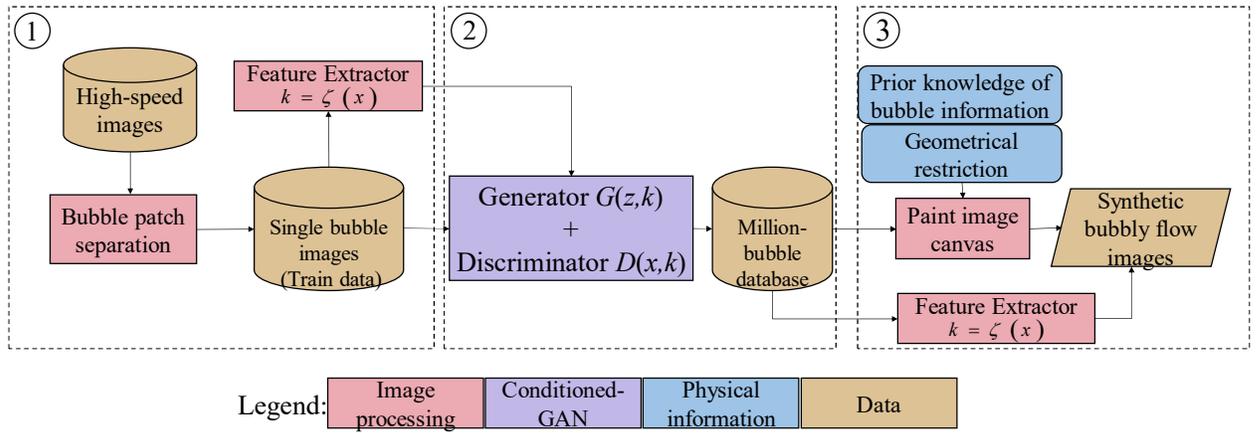

Fig. 2 The pipeline of the BubGAN algorithm. The BubGAN consists of three sections: (1) bubble training data extraction, (2) million-bubble database generation with conditional GAN and (3) bubbly flow image synthesis with given physical information. Various techniques and information used in BubGAN are shown in different colors as indicated by the legend.

### 3.1 Image processing technique

Various image processing tasks have been performed throughout the BubGAN workflow. To train the GAN algorithm, a sufficient amount of real bubble samples should be prepared. The bubbly flow images were first segmented into small image patches which either contain a single bubble or a number of bubbles which are connected together. For each image patch, three different algorithms, namely, watershed segmentation, bubble skeleton and adaptive threshold methods are used to determine whether this patch contains only a single bubble. The detailed procedures with sample intermediate results are shown in Fig. 3. Each algorithm will give a prediction of bubble number in the patch as denoted as $N_1$, $N_2$, $N_3$, respectively. Then the combined number prediction $N$ is assigned as the modal value $N_1$, $N_2$ and $N_3$. Once a patch is identified as a single bubble, it will proceed to visual checking step for results confirmation. In this study, 10000 single bubble image patches are extracted from the bubbly flow images.

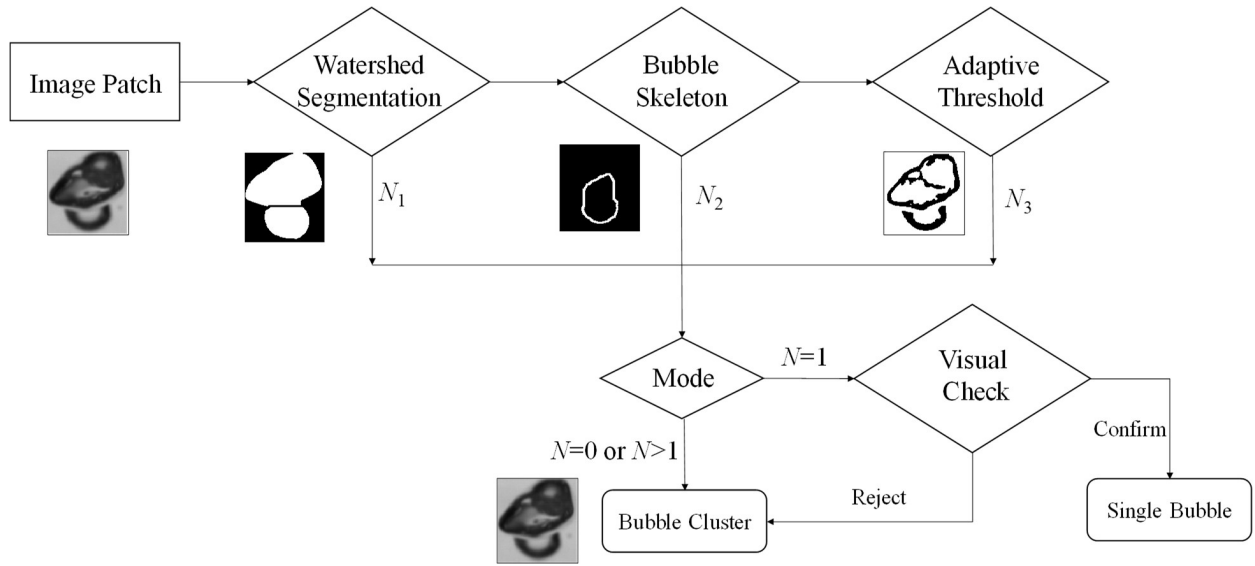

Fig. 3 The flowchart of identifying single bubble and bubble cluster patches. The process involves watershed segmentation, bubble skeleton and adaptive threshold methods for bubble number prediction in an image patch. Visual checking will be applied before a patch is finally classified as a single bubble. The figure also includes a sample image patch, which is classified as a bubble cluster.

After classification, all single bubble patches are further processed and normalized to serve as the training data for GAN. These steps are summarized graphically in Fig. 4. An example of a separated bubbly flow image is shown in Fig. 4 (a). The bubble cluster and single bubble patches are separated by the aforementioned algorithms. Those single bubbles are cropped by a square window for further processing. This will ensure that the aspect ratio of the bubble is preserved in the following normalization process. In these patches, other bubble fragments may appear as shown in Fig. 4 (b). To remove these fragments, the grayscale bubble image is converted to a binary image as shown in Fig. 4 (c). The fragment can be identified using the binary image and it is removed by filling the region with the background. The bubble mask and single bubble image after noise removal are shown in Fig. 4 (d) and Fig. 4 (e), respectively. Each separated single bubble patch is then normalized and rescaled to a size of 64 × 64 pixels for the GAN training in the next step.

In addition to training data preparation, image processing algorithm is also used in a later step for bubble feature extraction and bubbly flow image assembling. More details of the image processing techniques used in this paper can be found in the previous work reported by Fu and Liu [14,17].

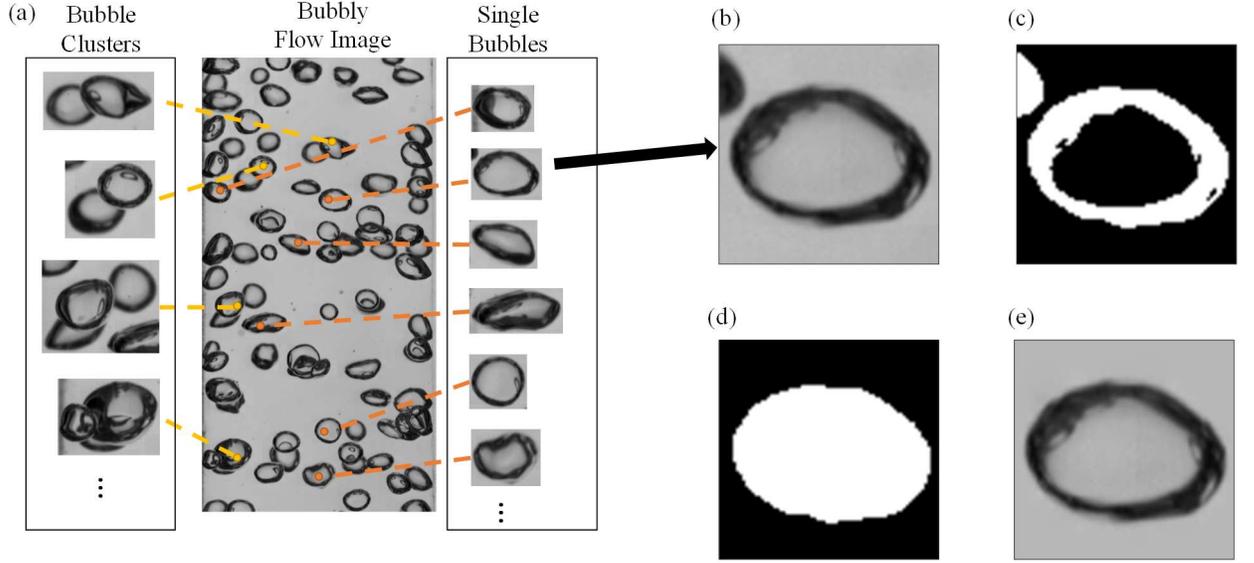

Fig. 4 Procedures of extracting single bubble patches for training data preparation (a) separate single bubbles and bubble clusters from recorded bubbly flow images, (b) extract square single bubble patch from image, (c) binarization of the grayscale image, (d) remove noise and create bubble mask (e) rescale the patch to 64 × 64 pixels.

### 3.2 Bubble feature extraction

Four parameters related to bubble geometry and topology are of primary interest in this study. These parameters are selected such that they can characterize typical bubbles during their transport processes in engineering applications. These parameters often appear in the engineering models or correlations. In addition, they will be used as one metric to assess the quality of the synthetic bubble images in a later section.

The first parameter is the bubble aspect ratio $E$, which is defined as:

$$E = \frac{b}{a}, \tag{1}$$

where $a$ and $b$ are the semi-major and semi-minor axes of the ellipse fitted on the bubble. The second parameter is the rotation angle $\varphi$ which is defined as the angle from the positive $x$ axis to the semi-major axis in an anti-clockwise direction. The third parameter $\Psi$ is called the circularity of the bubble, which is calculated as:

$$\Psi = \frac{4\pi A}{P^2}, \tag{2}$$

where $A$ represents the bubble projection area and $P$ stands for the perimeter of the bubble. If the bubble is close to a circle, $\Psi$ will be close to 1. If the bubble is elongated or has an irregular shape, $\Psi$ decreases towards 0. The fourth parameter, edge ratio $m$, is defined as the dark edge area $A_{edge}$ over the bubble projection area $A$ as:

$$m = \frac{A_{edge}}{A}. \tag{3}$$

This parameter can be used to represent bubble appearance. The darker edge corresponding to more total reflection, which can be caused by stretching bubble along the camera axis. In this case, the $m$ will have a large value. If the bubble is stretched in the two directions perpendicular to the camera axis, there will be less total reflection area and the *m* will be close to 0. Besides, the setup of the backlit illumination and bubble surface distortion can also be factors which affect value of *m*.

A summary of these parameters are listed in Table I. The range and unit of these parameters are given in the table. It should be noticed that the ranges in the table give the minimum/maximum possible values for reference. In real bubble images, the parameters may not reach the extreme value.

Table I. Description of extracted bubble feature parameters.

| Bubble parameter | Definition | Range | Unit |
|---|---|---|---|
| **Aspect ratio $E$** | $E = b/a$ | (0,1] | - |
| **Rotation angle $\varphi$** | Rotation of the fitted ellipse on the bubble image | (-π/2, π/2] | Rad |
| **Circularity $\Psi$** | $\Psi = 4\pi A / P^2$ | (0,1] | - |
| **Edge ratio $m$** | $m = A_{edge} / A$ | (0,1] | - |

### 3.3 Generative adversarial networks

The generative adversarial networks [29,30] is used for the generation of realistic single bubble images in this work. The GAN consists of two different neural networks, namely, the generative network defined by a function $G$ and the discriminator network defined by a function $D$. The discriminator function $D(x)$ takes an input of $x$, which represents the images used for training or testing. In this study, $x$ consists of 10000 training bubble patches extracted from the high speed images taken in bubbly flow condition. The discriminator network $D$ does the classification job for identifying the real bubbles and the synthetic bubbles generated by the generative network $G$. The generative network would generate images using function $G(z)$, where $z$ is the latent vector sampled from a prior normal distribution. The terms $\theta^{(D)}$ and $\theta^{(G)}$ are the parameters used in function $D$ and $G$. With the cross-entropy cost definition, the discriminator cost function is defined as:

$$L^{(D)}\left(\theta^{(D)},\theta^{(G)}\right) = -\frac{1}{2}\mathbb{E}_{x\sim p_x} \log D(x) - \frac{1}{2}\mathbb{E}_{z\sim p_z} \log\left(1-D(G(z))\right). \tag{4}$$

The cost function of the generator can be directly calculated as:

$$L^{(G)} = -L^{(D)}. \tag{5}$$

in a zero-sum game. To reach the Nash equilibrium of the game, one should optimize the parameters with the minimax of:

$$\arg\min_{\theta^{(G)}}\max_{\theta^{(D)}} L^{(G)}\left(\theta^{(D)},\theta^{(G)}\right). \tag{6}$$

In this work, the deep convolutional generative adversarial network [30] is adapted for bubble image generation due to its improved quality and sharpness in image generation. To have better control of the generated bubble images, both the generative and discriminator networks are conditioned on the bubble features for training. The detailed structure of the proposed conditional GAN is illustrated in Fig. 5. As shown in the figure, the input of a generative network consists of a bubble feature vector $k \in \mathbb{R}^4$ and a random latent noise vector $z \in \mathbb{R}^{100}$. The bubble feature vector $k$ contains four bubble parameters as summarized in Table I:

$$k = \zeta(x) = [E, \varphi, \Psi, m]. \tag{7}$$

where $\zeta(x)$ represents the feature extractor function based on the image processing algorithm.

After convolution, the feature vector is projected to 512 dimensions and concatenated in depth to both the Generative network $G$ and the discriminator network $D$. With the feedback given by the discriminator, the generator $G$ can finally learn to generate the realistic bubble images with its feature given by the feature vector $k$. The output bubble image from $G$ has a dimension of 64 × 64 × 3.

In the training process, the conditional GAN should be able to achieve two objectives. For the generator $G$, it should be able to generate bubbles with features defined by input feature vector $k$. For the discriminator $D$, it should be able to identify whether the feature vector $k$ matches the input bubble along with identifying the authenticity of the bubble image. To reinforce the connection between a bubble and its feature vector, three types of false pairs are designed for training the discriminator $D$. The first is using a fake image $\hat{x}$ with its corresponding input feature vector $k_2$. The second pair is the fake image $\hat{x}$ paired with an incorrect bubble feature vector $k_3$. The third pair is a real bubble image $x$ paired with an incorrect bubble feature vector $k_2$, which is used for the synthetic image generating. Then the loss function for the discriminator can be updated as:

$$L^{(D)} = -\frac{1}{2}\mathbb{E}_{x \sim p_x} y - \frac{1}{2}\mathbb{E}_{z \sim p_z}\left[\frac{1}{3}\log(1-\hat{y}_1) + \frac{1}{3}\log(1-\hat{y}_2) + \frac{1}{3}\log(1-\hat{y}_3)\right], \tag{8}$$

where $\hat{y}_1 = D(\hat{x}, k_2), \hat{y}_2 = D(\hat{x}, k_3), \hat{y}_3 = D(x, k_2)$, and $y = D(x, k_1)$ represents the score given by the discriminator with the input of real image $x$ along with its feature vector $k_1$. The details of training the networks are shown in **Algorithm 1**.

With the trained GAN conditioned on bubble features, a million bubbles are generated to create a database for bubbly flow image synthesis. These generated bubbles are preprocessed to get their companion feature vector $k$. The bubble feature information is associated with each bubble, which will allow a quick search of the desired bubble image given a feature vector input. This database will improve the efficiency of generating synthetic bubbly flow images to be discussed in the coming steps.

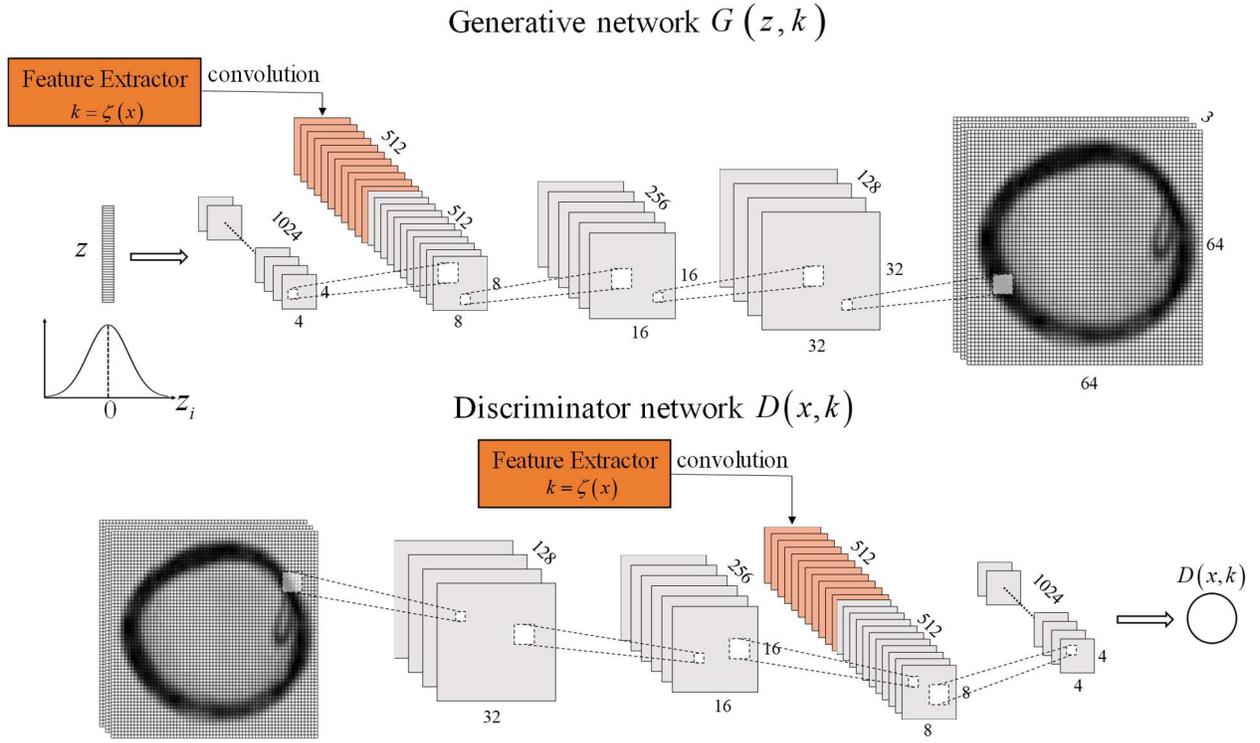

Fig. 5 The structure of proposed generative adversarial networks (GAN) conditioned on the bubble features.

**Algorithm 1**
*for* the number of training epochs *do*
*for* the number of minibatchs *do*

1. $k_1 = \zeta(x)$ **Extract matching bubble feature vectors from minibatch images**
2. $k_2, k_3$ **Draw random bubble feature vectors from the training images**
3. $\hat{x} = G(z, k_2)$ **Generate fake images with the Generator network $G$**
4. $y = D(x, k_1)$ **Calculate a score for real image with matching feature vectors**
5. $\hat{y}_1 = D(\hat{x}, k_2), \hat{y}_2 = D(\hat{x}, k_3), \hat{y}_3 = D(x, k_2)$ **Generate fake inputs for the discriminator**
6. **Minimize** $L^{(D)} = -\frac{1}{2}\mathbb{E}_{x \sim p_x} y - \frac{1}{2}\mathbb{E}_{z \sim p_z}\left[\frac{1}{3}\log(1-\hat{y}_1) + \frac{1}{3}\log(1-\hat{y}_2) + \frac{1}{3}\log(1-\hat{y}_3)\right]$
7. **Maximize** $L^{(G)} = \frac{1}{2}\mathbb{E}_{x \sim p_x}(1 - D(\hat{x}, k_2))$ **or equivalently minimize**

$-\frac{1}{2}\mathbb{E}_{z \sim p_z}(D(\hat{x}, k_2))$ **in practice**

**end** *for*
**end** *for*

## 3.4 Physical information and geometrical restriction

With given physical information, such as, bubble size and number density distribution, synthetic bubbly flow images can be assembled using the million-bubble database. In addition, geometrical restriction should be applied on bubbles to prevent them from going outside the physical boundary. The flowchart of generating upward bubbly flow images are shown in Fig. 6. In this process, the flow region and imaging area should be given as a prior. Bubble size distribution and void fraction distribution in the lateral direction together determine the averaged bubble occurrence frequency and size in the images. With this information, a bubble list can be created with known semi-axes ($a$, $b$), location ($x$, $y$, $z$), and aspect ratio $E$, etc. Based on this list, the million-bubble database will be searched to select the bubbles with matching parameters. Finally, these bubbles will be painted onto the image canvas according to their center locations ($x$, $y$, $z$). Since the million-bubble database has a standard image size of 64×64 pixels, the bubble image should be rescaled to the desired size specified by the bubble list for painting. As shown in Fig. 6 (d), when painting the bubble onto the image canvas, the flow boundary restriction should be enforced. In another word, any part of a bubble should not exceed the physical boundary of the flow channel. The numbered bubbles in Fig. 6 (d) are three typical scenarios when a bubble is partially outside the imaging area, where the left and right boundary lines also coincide with the physical boundary of the vertical flow channel. The first bubble will be shifted left to make sure the entire bubble is inside the physical boundary. For the second bubble, the lower part is partially outside the image boundary. However, the bubble is still within the physical boundary of the flow channel. Therefore, this part will just be cropped in painting. For the third bubble on the top-left corner, it will be shifted to the left first, then the upper part will be cropped to meet the geometrical restriction requirement.

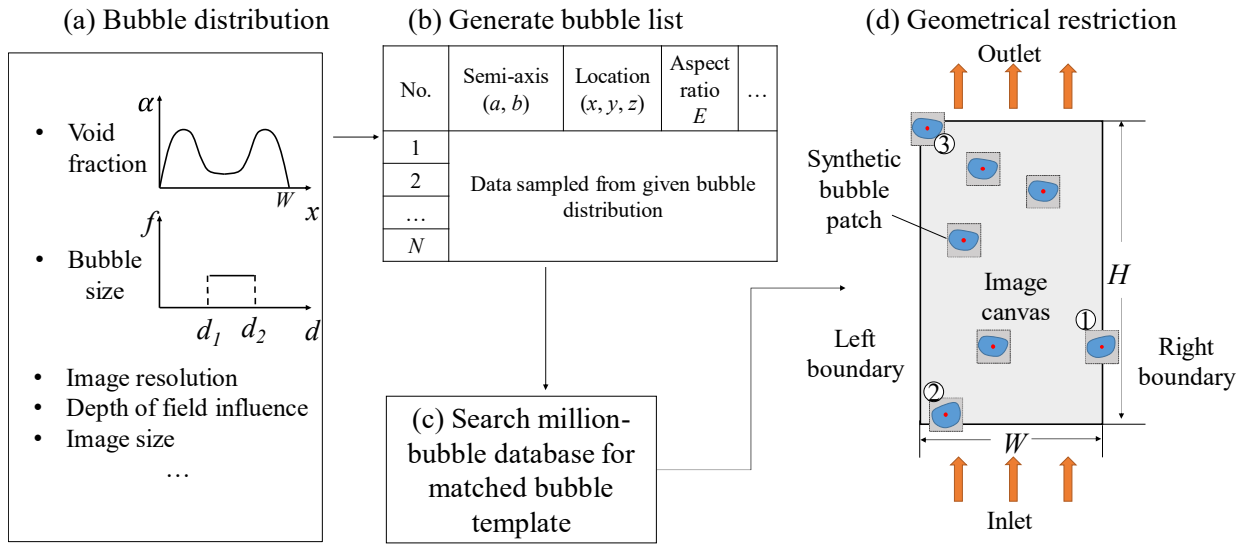

Fig. 6 Flowchart of generating synthetic bubbly flow images with the million-bubble database. Additional physical information such as bubble size and location distribution is given as priors. Geometrical restriction is applied to prevent bubbles from going outside the physical boundary of flow channel.

## 4 Results and discussions
### 4.1 Quantitatively assessment of BubGAN

It becomes feasible to generate realistic bubbly flow images using the proposed BubGAN algorithm. To assess the diversity of the million-bubble database, two sets of synthetic bubbles sampled from the database are shown in Fig. 7. The first image collection shows bubbles with various aspect ratios and rotation angles. The second image collection shows bubbles with various edge ratios and circularities. It is seen that the million-bubble database contains bubbles with a large variance in bubble features. The abundant bubble shapes and appearances should be able to synthesize bubbles appeared in most bubbly flow conditions.

As discussed in Section 3, the GAN is conditioned on bubble features. The generator $G(x, k)$ can generate synthetic bubbles according to the given feature vector $k$. To check whether the generator $G$ can reproduce correct bubble images, three tests are carried out with different feature vectors. The results are shown in Fig. 8. In the first test, bubble rotation angle is fixed at $\varphi_{Cond} = 0.53$ in the input vector $k$. Then a total of 64 bubbles are generated using the trained generator $G$. Among these generated bubbles, the averaged rotation angle is $\bar{\varphi}_{Gen} = 0.52$ with a relative error of 2.38%. The second and the third histograms show bubble aspect ratio and edge ratio distribution with given feature input of $E_{Cond} = 0.61$ and $m_{Cond} = 0.78$, respectively. The results show a relative error less than 2% in both cases.

The above tests use statistical data to access the quality of the BubGAN, which is conditioned on bubble features. An exhaustive evaluation of the BubGAN performance on generating bubbles is shown in Fig. 9. In the evaluation, synthetic bubbles are generated by varying a specific parameter at a time. For example, Fig. 9 (a) shows an example of generating bubbles by varying rotation angle $\varphi_{Cond}$ from negative to positive, which is shown along the $x$ axis. The $y$ axis represents the result $\bar{\varphi}_{Gen}$ obtained from the generated bubbles. Each data point in the figure represents the value averaged over 100 sample images. The root-mean-square error (RMSE) is calculated using all 10 data points for each parameter to assess the overall performance of the BubGAN. For all four feature parameters, the RMSE is less than 2.5%. This indicates that the generator $G$ is capable of reproducing desired bubbles over a wide range of conditions.

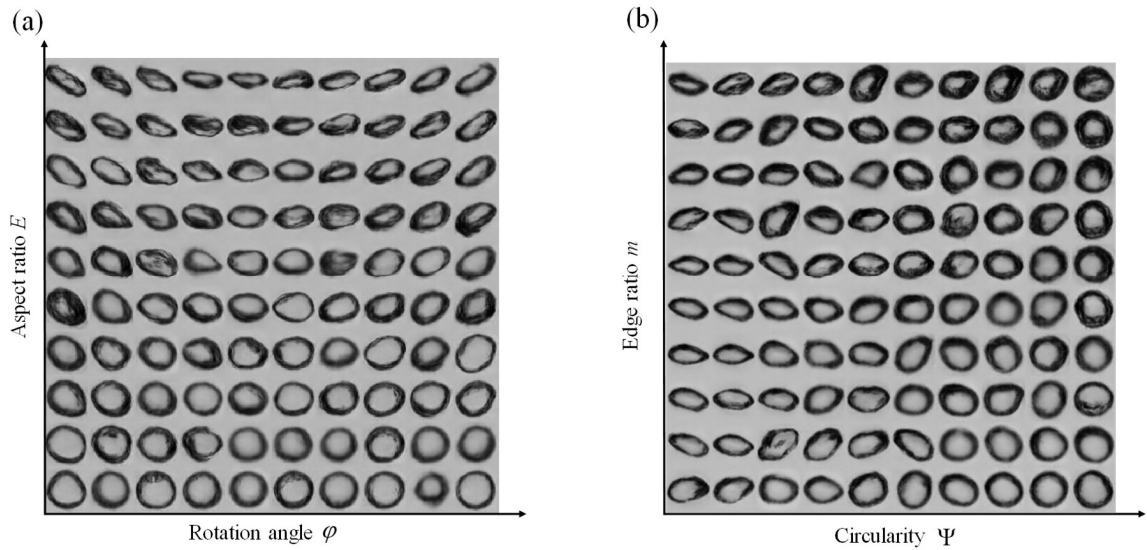

Fig. 7 Single bubble image samples selected from the million-bubble database with various features of (a) aspect ratio $E$ and rotation angle, (b) edge ratio $m$ and circularity .

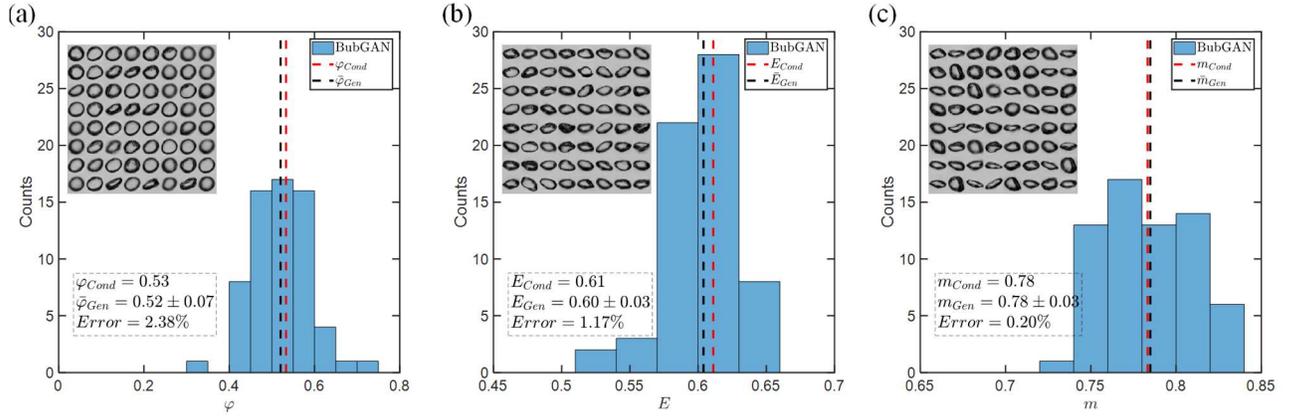

Fig. 8 A quantitative study of synthetic bubble generation conditioned on given bubble parameters of (a) rotation angle $\varphi_{Cond} = 0.53$, (b) aspect ratio $E_{Cond} = 0.61$ and (c) edge ratio $m = 0.78$.

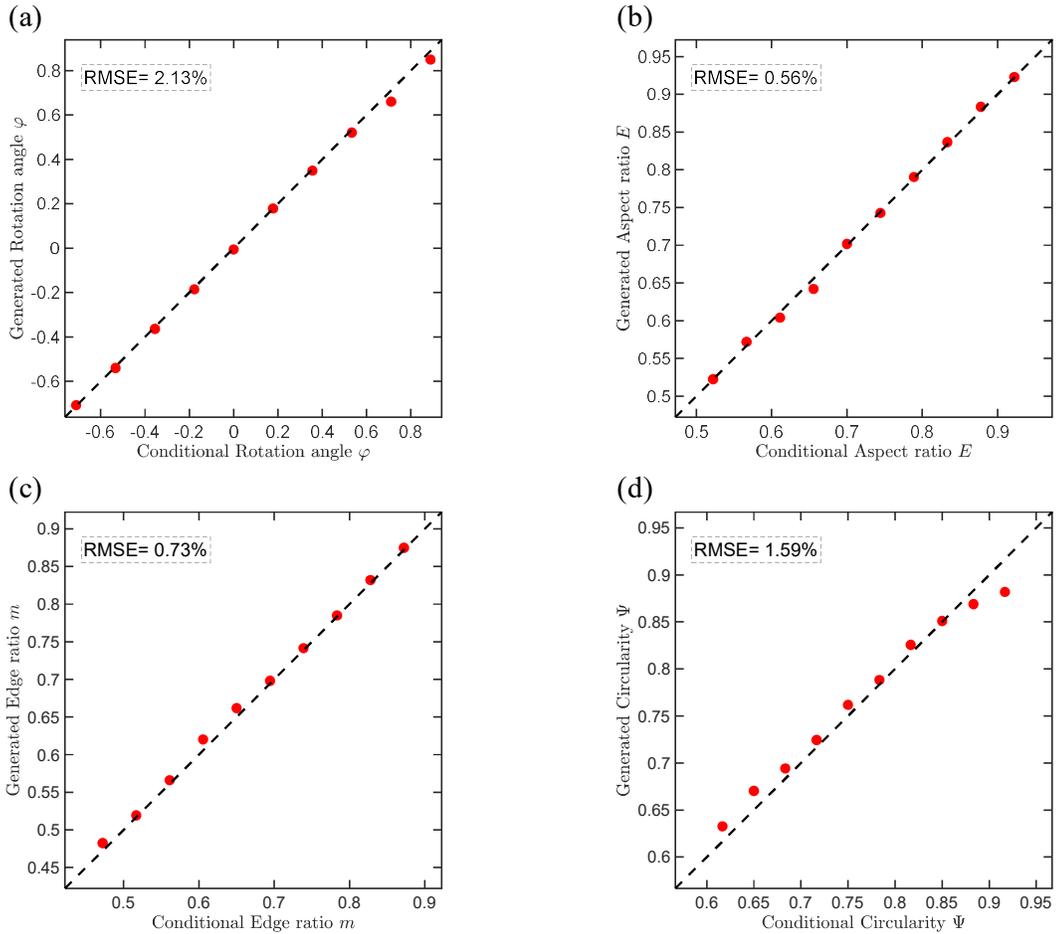

Fig. 9 Statistical evaluation of the conditioned GAN on generating the bubbles on given bubble parameters of (a) rotation angle, (b) aspect ratio, (c) edge ratio and (d) circularity. The root-mean-square error calculated from ten data points in each plot. Each data points are the averaged value calculated from 100 generated bubble samples with a given fixed bubble feature value.

It should be noticed that the input bubble feature vector $k$ should be sampled from the training dataset. This is because the conditional GAN learns bubble features in the manifold of bubble feature space defined by the training dataset. A random combination of four bubble feature components can result in unphysical feature vector $k$, which will downgrade the image quality. The correlations among different feature parameters of the training dataset are shown in Fig. 10. As shown in the figure, different parameters have a certain degree of correlation with other parameters. Therefore, the bubble feature vector $k$ has a degree of freedom less than 4. In the generation of synthetic bubbles, the four feature components cannot be specified randomly without considering the potential correlations among them. To augment the feature vector space, interpolation of two feature vectors can be used to add variation of bubble shapes and appearances. This method has been proven to work well in deep representation mixing [31] and text to image synthesis [32]. In this work, a new bubble feature vector $k$ is interpolated by combining two existing feature vectors ($k_i$ and $k_j$) randomly sampled from the training dataset as:

$$k = \beta k_i + (1-\beta)k_j, \tag{9}$$

where $\beta$ can be a value from 0 to 1. With this interpolation, the newly generated bubbles show good quality and the feature vectors can match the input. An example is shown in Fig. 11 with five spider plots by varying $\beta$ from 0 to 1. The blue lines in the figure represent the interpolated input feature vector, and the red dashed lines represent the actual values of generated bubbles. The results show a very good match between these two, which indicates that the trained generator $G$ can generate bubbles with an arbitrary feature vector within the convex hull formed by the training dataset.

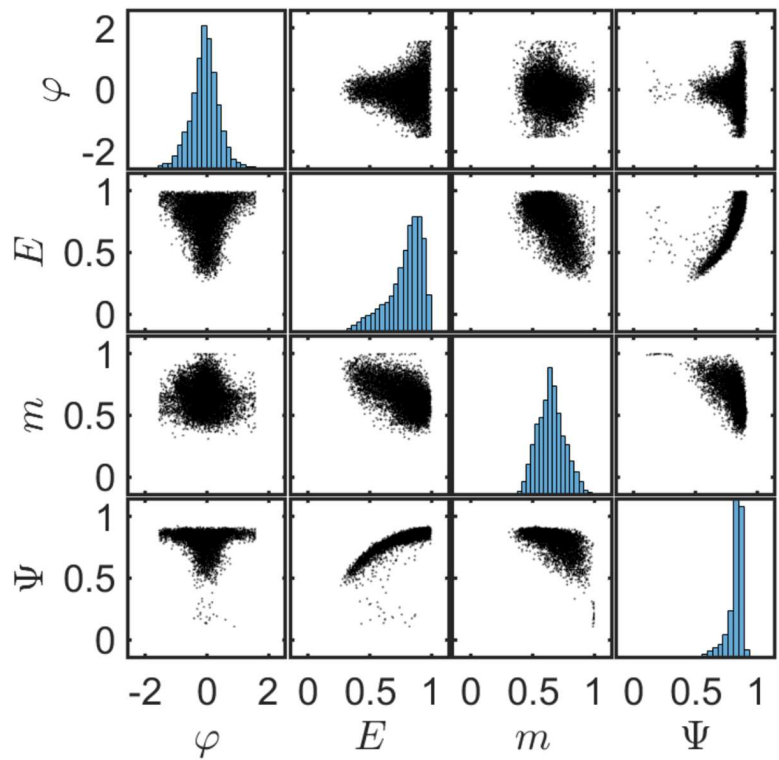

Fig. 10 A matrix plot of the correlations between four bubble features acquired from the training data.

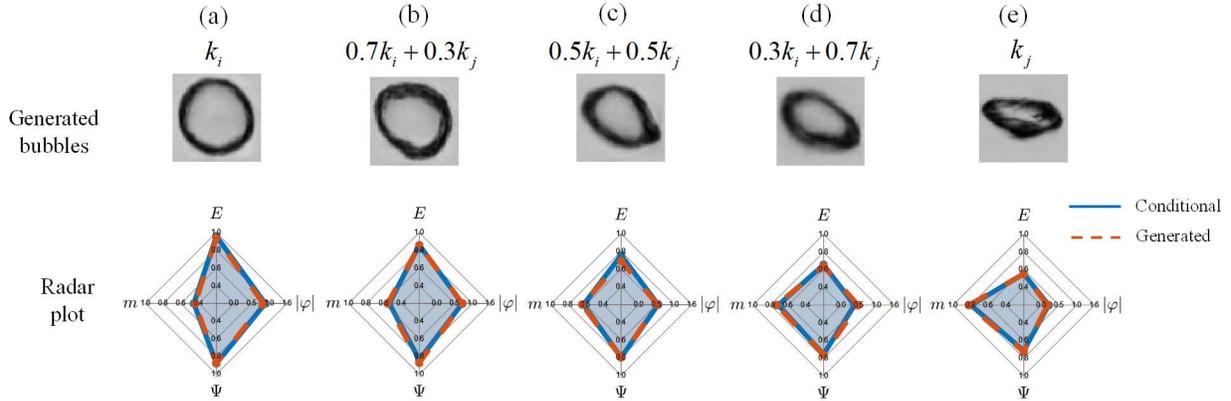

Fig. 11 Example of interpolating two bubble feature vectors $k_i$ and $k_j$ to generate bubbles outside the training dataset manifold. The five examples show the generated bubbles using the feature vector of (1) $k_i$, (2) $0.7k_i+0.3k_j$, (3) $0.5k_i+0.5k_j$, (4) $0.3k_i+0.7k_j$ and (5) $k_j$.

### 4.2 Bubbly Flow Synthesis

To demonstrate the performance of the proposed BubGAN, a visual comparison of real and synthetic images are shown in Fig. 12 for typical bubbly flows. The real image captured by high speed camera is shown in Fig. 12 (a). Fig. 12 (b) and (c) show synthetic images generated using the conventional CCA method and BubGAN, respectively. It can be seen that the synthetic image generated by the CCA method, as shown in Fig. 12 (b), shows a noticeable difference compared to the real image. All the bubbles in the figure share a similar appearance and they do not present shape and surface distortions as appeared in the real image. The synthetic image generated by the BubGAN, as shown in Fig. 12 (c), shows a significant improvement in both shape variations and visual appearances. In the following, more examples are given by generating bubbly flow images in different flow channels.

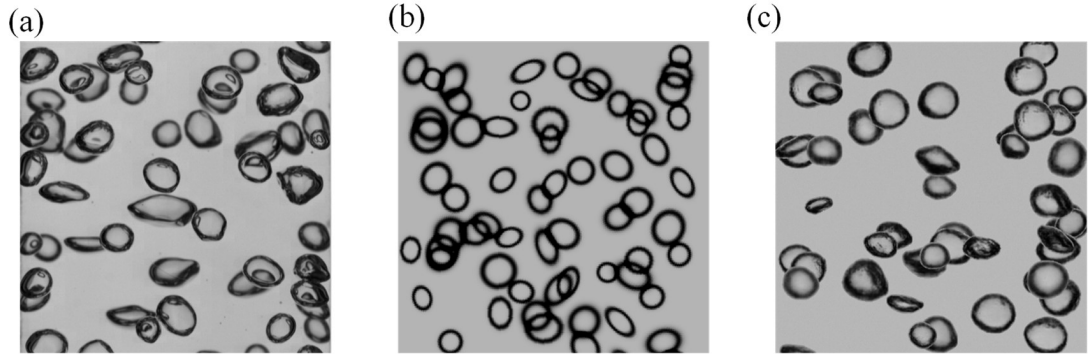

Fig. 12 Comparison of (a) the real bubbly flow images with the synthetic bubbly flow images generated based on (b) CCA model and (c) the BubGAN algorithm.

#### 4.2.1 30 mm × 10 mm Rectangular Channel

For the rectangular channel with a cross section of 30 mm × 10 mm, air and water are injected uniformly at the inlet with the two-phase injector described previously in Fig. 1. Four test conditions are shown in Fig. 13, all with a superficial gas velocity fixed at $j_g = 0.1$ m/s. The superficial liquid velocity for Run 1-4 are $j_f = 0.5$ m/s, 1.0 m/s, 1.28 m/s and 2.12 m/s, respectively. High speed images are recorded at 1000 frames per second with a resolution of around 25 pixels per mm. The gas phase void fraction decreases with superficial liquid velocity $j_f$ as $j_g$ is a constant, whereas turbulence intensity increases with the increase of $j_f$. Under these varying void fraction and turbulence conditions, bubble size and number density decrease from Run 1 to Run 4. As can be seen in Fig. 13, images generated by BubGAN can faithfully reflect the trends of bubble size and number density changes. The bubble shape is not limited to circular or elliptical shapes as compared to the CCA method.

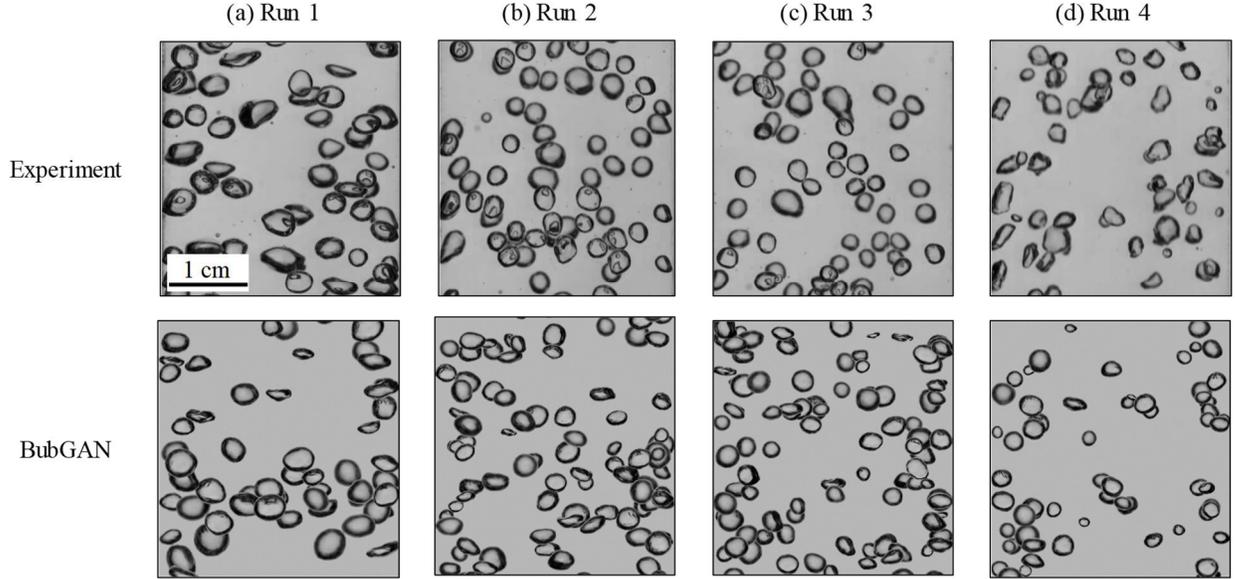

Fig. 13 Comparison of the experiment images and the synthetic images generated by BubGAN the 30 mm × 10 mm rectangular test section. The comparison includes the flow conditions of (a) $j_f$ = 0.5 m/s, $j_g$ = 0.1 m/s, (b) $j_f$ = 1.0 m/s, $j_g$ = 0.1 m/s, (c) $j_f$ = 1.28 m/s, $j_g$ = 0.1 m/s, (d) $j_f$ = 2.12 m/s, $j_g$ = 0.1 m/s.

4.2.2   200 mm × 10 mm Rectangular Channel

Another example of synthetic bubbly flow image is given for a rectangular test loop with a cross section size of 200 mm × 10 mm. The test section has the ability to control the gas flow rate along the 200 mm width direction to form non-uniform bubble distributions. More details of the test facility can be found in previous works [33,34].

In this test channel, three gas injection modes are shown in Fig. 14. The experimental images are recorded at the height of $z/D_h$ = 10, where $D_h$ is the hydraulic diameter which is 19.05 mm for the test channel. The recorded images have a resolution of 4.77 pixels per mm. In Fig. 14 (a), gas is injected from the center region of the flow channel. The bubbles are mainly concentrated in the center region, and no bubble exists in the regions near both side walls. In Fig. 14 (b), the double peak mode is formed by injecting gas near the left and right walls. The bubble number density decreases when moving towards the center region. In the third example as shown in Fig. 14 (c), the gas is injected near the left wall to form a single side peak pattern. As shown in the figure, the BubGAN can generate all three bubbly flow patterns accurately. The spatial distributions of bubbles in different inlet conditions are correctly reproduced in the synthetic images.

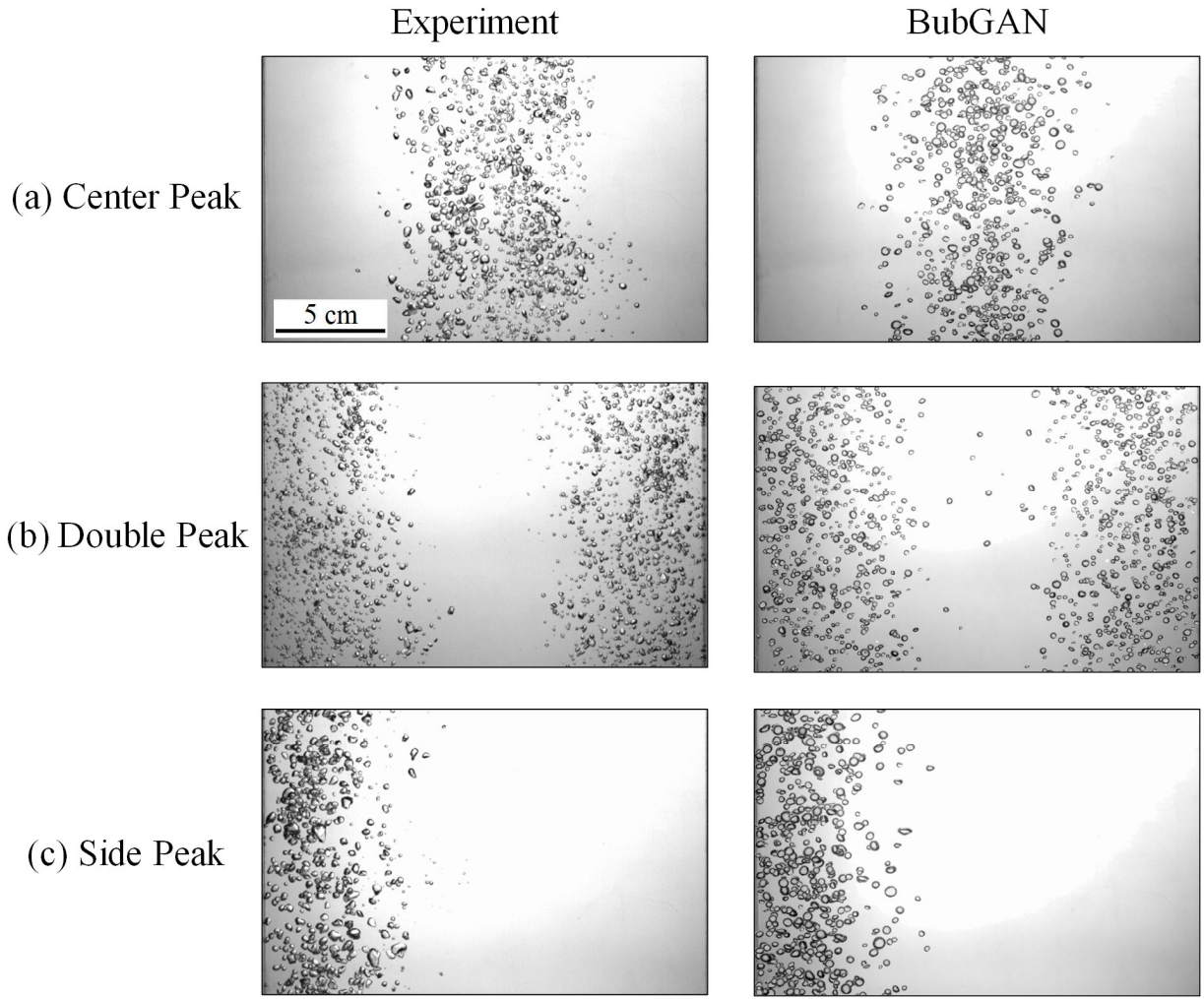

Fig. 14 Comparison of experimental images and synthetic images generated by BubGAN for the 200 mm × 10 mm rectangular test section. The generated images include three inlet bubble distributions (a) center peak, (b) double peak and (d) side peak.

### 4.3  Bubble Image Labeling

Detailed bubble label information is also available when synthetic images are generated using the BubGAN algorithm. Examples of bubble labeling are demonstrated in Fig. 15. The figure shows a synthetic bubbly flow image along with projection area, aspect ratio $E$ and rotation angle $\varphi$ labeled for each individual bubbles. Such information is automatically generated by the BubGAN algorithm without the need of any manual processing. These well-labeled images can be used to benchmark existing image processing algorithms [14, 17]. The red dots in Fig. 15 (b)-(d) represent the center of each bubble. This can be used to generate a bubble density map along with the synthetic bubble images. With this information, various machine learning algorithms [3–5,35,36] can be trained for bubble counting without additional cost for bubble labeling. The realistic images can also reduce the errors introduced by the discrepancy between synthetic and real images.

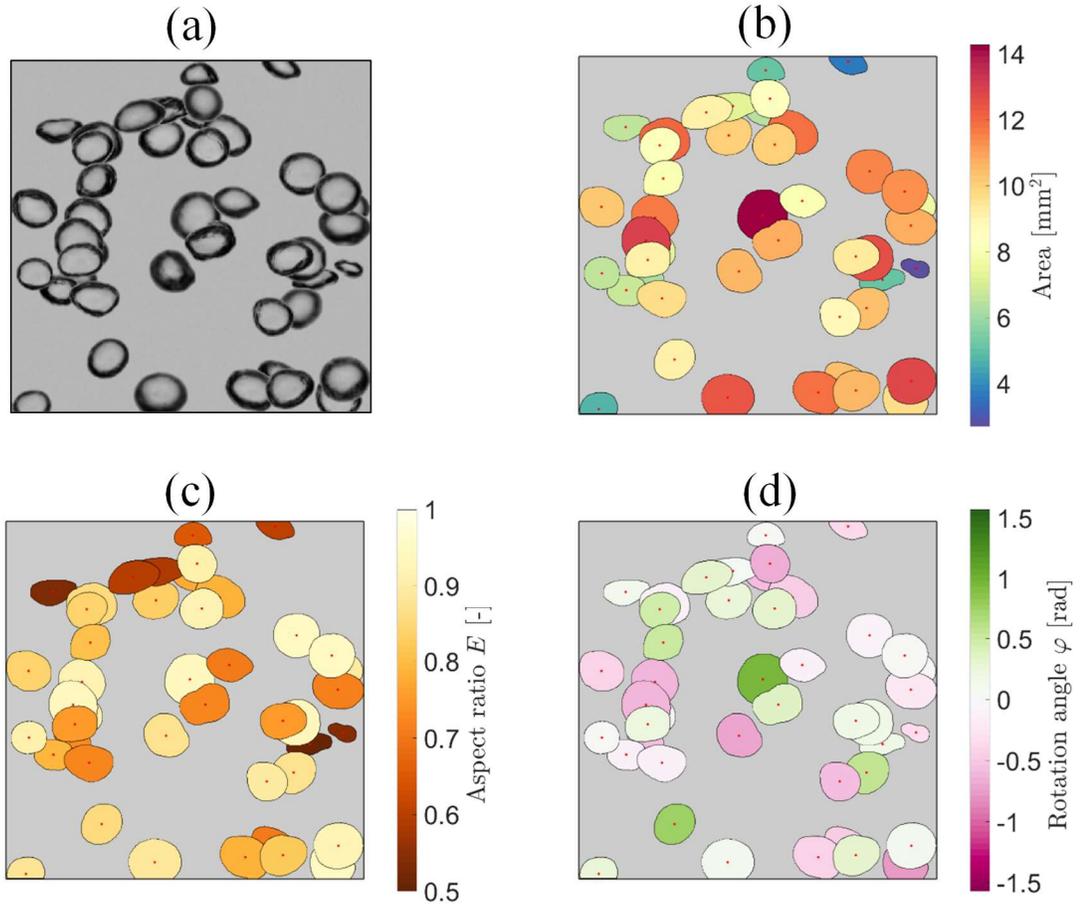

Fig. 15 Generation of a uniform distributed bubbly flow image with a million-bubble database with labeled bubble information, (a) synthetic image, (b) projection area, (c) aspect ratio and (d) rotation angle.

## 5   Conclusion

This paper proposed a BubGAN algorithm for bubbly flow image synthesis. The BubGAN combines image processing algorithms and the conditional-GAN to achieve automatic bubbly flow image generation and labeling. The generated flow images show a significant improvement compared to the conventional bubble models. This can help to assess the accuracy of existing image processing algorithms and to guide future algorithm development. The deep learning based object counting algorithms can also benefit from the BubGAN by greatly reducing the labeling cost. Four bubble feature parameters, namely, rotation angle, aspect ratio, edge ratio and circularity are integrated into the BubGAN algorithm. Therefore, one can generate a wide range of bubbles with desired bubble features. The quantitative assessments in Fig. 8 and Fig. 9 demonstrate the reliability and accuracy of the BubGAN in generating bubbles with given parameters. A tool for bubbly flow image generation is also developed based on a million-bubble database pre-generated by the BubGAN. The tool is capable of generating spatially uniform, as well as non-uniform bubbly flow images with detailed labeling information. Therefore, algorithm benchmarking and related deep learning algorithm training can be carried out without the additional cost of labeling.

## 6 Acknowledgment

We would like to acknowledge Yufeng Ma, Srijan Sengupta and Ahmed Ibrahim for their helpful discussions on this project. We would like to thank Sung-Ho Bae for his discussion on the implementation of bubble features in the conditional GAN.